%% file: IJCNN_CS-VAE.tex
\documentclass[conference,table]{IEEEtran}
\IEEEoverridecommandlockouts
\usepackage{cite}
\usepackage{amsmath,amssymb,amsfonts}
\usepackage{algorithmic}
\usepackage{graphicx}
\usepackage{textcomp}
\usepackage{xcolor}
\usepackage{lipsum}
\usepackage{verbatim}
\usepackage{amsmath}
\usepackage{relsize}
\usepackage{xfrac}
\usepackage{multirow}
\usepackage{float}
\usepackage{mathtools}
\usepackage{soul}

\usepackage[sort&compress,numbers,square]{natbib}

\def\BibTeX{{\rm B\kern-.05em{\sc i\kern-.025em b}\kern-.08em
    T\kern-.1667em\lower.7ex\hbox{E}\kern-.125emX}}
\begin{document}

\title{Class-Specific Variational Auto-Encoder\\for Content-Based Image Retrieval\\
}

\author{\IEEEauthorblockN{Mehdi Rafiei and Alexandros Iosifidis}
\IEEEauthorblockA{\textit{Department of Electrical and Computer Engineering} \\
\textit{Aarhus University, Aarhus, Denmark} \\
\{rafiei,ai\}@ece.au.dk}}
\vspace{-10mm}

\maketitle

\begin{abstract}
Using a discriminative representation obtained by supervised deep learning methods showed promising results on diverse Content-Based Image Retrieval (CBIR) problems. However, existing methods exploiting labels during training try to discriminate all available classes, which is not ideal in cases where the retrieval problem focuses on a class of interest. In this paper, we propose a regularized loss for Variational Auto-Encoders (VAEs) forcing the model to focus on a given class of interest. As a result, the model learns to discriminate the data belonging to the class of interest from any other possibility, making the learnt latent space of the VAE suitable for class-specific retrieval tasks. The proposed Class-Specific Variational Auto-Encoder (CS-VAE) is evaluated on three public and one custom datasets, and its performance is compared with that of three related VAE-based methods. Experimental results show that the proposed method outperforms its competition in both in-domain and out-of-domain retrieval problems.
\end{abstract}


\section{Introduction}
\label{sec:Introduction}
\IEEEPARstart{C}{ontent}-Based Image Retrieval (CBIR) is the task of searching for images with similar content to that of a query image. Available methods for doing so can be broadly categorized into those based on feature engineering and those relying on deep learning models \citep{chen2022deep}. Methods belonging to the first category adopt in their first stage explicit feature extraction techniques (e.g. based on color \citep{alam2020dynamic}, texture \citep{liu2013content}, and shape \citep{singha2012content}), along with keypoint-based image descriptions and representations (e.g. the Scale Invariant Feature Transform (SIFT) \citep{lowe2004distinctive} combined with Bag of Words (BoW) model \citep{sivic2003video}, or the Fisher Vector (FV) representation \cite{uchida2016image}).  

In recent years, deep learning models outperformed the more traditional methods in diverse computer vision tasks, including CBIR, with convolutional layers being widely used for the feature extraction \citep{pan2022generating, hamreras2021dynamic, pang2018deep, radenovic2018fine}. Such data-driven feature extractors can be part of Convolutional Neural Network (CNN) models which are either pre-trained on large image datasets \citep{pang2018deep}, e.g., like the ImageNet \citep{deng2009imagenet}, or be fine-tuned on a target dataset usually leading to improved retrieval performance \citep{radenovic2018fine}.

A type of deep learning models which is well-suited for CBIR is Variational Auto-Encoders (VAEs). Such models are trained to reconstruct their input, and the representation learnt in the latent space (output of the Encoder) can be used for CBIR. Traditional VAEs have limitations when applied in retrieval tasks, as they are not trained to discriminate between different classes forming the retrieval problem. To address this issue, one can train or fine-tune the model based on other reconstruction losses. However, a balance needs to be kept between generative and discriminative properties of the adopted loss as highly discriminative data representations can lead to losing valuable information which is important for image retrieval, especially when retrieving images based on queries belonging to classes outside of those appearing in the training (out-of-domain retrieval) \citep{passalis2016entropy}. To address this problem, a regularized discriminative deep VAE method was proposed in \citep{passalis2020variance} that models the latent generative factors for each of the training classes. Although this method shows good results in in-domain and out-of-domain image retrieval tasks, it is well-suited for multi-class retrieval problems and has limitations when it is applied to class-specific retrieval problems.

In this paper, we consider class-specific retrieval problems where one wants to retrieve images from a database based on a query image which either belongs to a class of interest, or not. In such a problem, the class of interest (called positive class) is usually well populated and all other images belong to multiple classes (forming the negative class of the class-specific problem) which are underpopulated, or their labels may not be available during the training phase. An example application which falls within this problem description is that of face retrieval where a specific person (e.g., the user of a system) is well-represented in the database while the rest of the facial images in the database can belong to other individuals with only a few (or even one) images. Following a multi-class formulation similar to that of \citep{passalis2016entropy} would lead to a binary problem where the negative class is modeled to be homogeneous, which generates problems as negative class images can exhibit very large variations.

\begin{figure*}[hbt]
\centering
\includegraphics[width=0.65\textwidth]{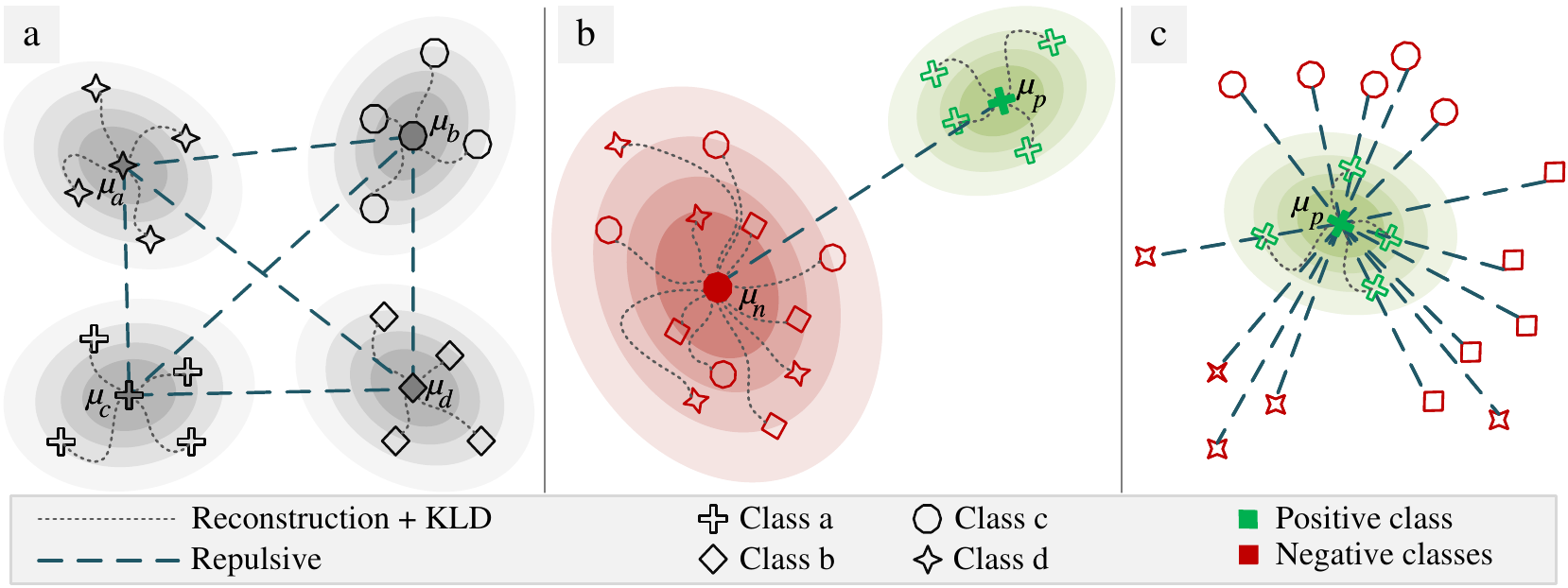}
\caption{Forces caused by reconstruction, KLD, and repulsive losses in a) RD-VAE, b) binary RD-VAE, and c) CS-VAE.}
\label{Losses}
\end{figure*}

To address this, we propose a training loss to train class-specific deep VAEs for CBIR. Instead of learning image representations capable to discriminate between different classes, the VAE is trained to discriminate between the class of interest (positive class) and any other available possibility (samples belonging to the negative class). This strategy encourages the model to learn data representations which leads to a homogeneous positive class in the latent space of the VAE and discriminates this class from the data belonging to any other classes forming the negative class. This would also help the model to discriminate the positive class from other classes unseen during training (out-of-domain retrieval). 
To demonstrate the performance of the proposed method over different data types, it is extensively evaluated on three public and one custom datasets. In addition, for each dataset, two evaluation scenarios (namely in- and out-of-domain retrieval) are considered. The performance of the method is compared with three other VAE-based methods.

The rest of this paper is structured as follows. In Section \ref{sec:related}, related work on VAE-based image retrieval is briefly discussed. The proposed method is described in \ref{sec:csvae}. Experimental evaluations are presented in Section \ref{sec:Imp}. Finally, the paper is concluded in Section \ref{sec:conclusion}.

\section{Related work}\label{sec:related}
Several supervised and unsupervised VAE methods, such as VAE \citep{kingma2013auto} and regularized discriminative VAE (RD-VAE) \citep{passalis2020variance}, have been proposed for CBIR tasks. However, these methods have limitations in class-specific CBIR applications. A VAE is commonly formed by an Encoder which receives an image as input and outputs a representation of that image in a (usually lower-dimensional) latent space, and a Decoder which receives this image representation as input and tries to reconstruct the input to the VAE image on its output. The parameters of both the Encoder and the Decoder are jointly optimized to preserve the maximum information when encoding and to have the minimum reconstruction error when decoding. By considering $x_i, \:i=1,\dots,N$ as the input images in the training set, this optimization is achieved by using the reconstruction loss and the Kullback–Leibler divergence:
\begin{equation}
L_{vae}\!=\!MSE_{loss}\!+\!\alpha_{\rm KL}KLD_{unsup},\label{L_vae}
\end{equation}
\begin{equation}
KLD_{unsup}\!=\!\mathlarger{\sum_{x_i}}{\left({(\,x_i\!-\!\mu)\,}\!^2\!+\!\sigma^2\!-\!\log(\,\sigma)\,\!-\!1\right)}.\label{KLD_uns}
\end{equation}
The Kullback–Leibler divergence term ($KLD_{unsup}$) forces the data representations in the latent space to form a Gaussian distribution having mean $\mu$ and variance $\sigma$. Using such an unsupervised learning process to train the VAE determines a Gaussian distribution that can be considered as the data distribution of all the training data (irrespectively of which class each training image may belong to) in the input of the Decoder. Thus, applying CBIR using image representations coming from the latent space of a VAE usually leads to low performance.  

To solve this limitation, a VAE-based regularized discriminative deep metric learning method (RD-VAE) was proposed in \citep{passalis2020variance}. This method modifies the training loss of the VAE such that samples belonging to the same class form homogeneous clusters which are well-separated by clusters corresponding to other classes. To do that, $KLD_{unsup}$ is replaced with a supervised KLD, and a repulsive term ($rep_{loss}$) is added to the loss function:
\begin{equation}
L_{\operatorname{rd-vae}}\!=\!MSE_{loss}\!+\!\alpha_{\rm KL}KLD_{sup}\!+\!rep_{loss},\label{L_rd}
\end{equation}
\begin{equation}
KLD_{sup}\!=\!\\
\mathlarger{\sum_{x_i}}{\left({(\,x_i\!-\!\mu_{l_i})\,}\!^2\!+\!\sigma_{l_i}\!^2\!-\!\log(\,\sigma_{l_i})\,\!-\!1\right)},\label{KLD_s}
\end{equation}
\begin{equation}
rep_{loss}\!=\!
\sfrac{1}{\rho}\mathlarger{\sum_{x_i}}\mathlarger{\sum_{x_j\ne x_i}}{\max\left(0,\rho \!-\!\parallel \mu_{l_i}\!-\!\mu_{l_j} \parallel_{2}^{2}\right)\!^2},\label{Rep}
\end{equation}
where $l_i$ denotes the class label of $x_i$, $KLD_{sup}$ determines a Gaussian distribution for each class, and $rep_{loss}$ forces the means of different class distributions away to be in a minimum distance of $\rho$ from each other. Those data representations are also constrained by the $MSE_{loss}$ loss, meaning that they need to preserve adequate input information in order to reconstruct the input image in the output of the Decoder. Figure \ref{Losses}-a shows a schematic 2D representation of RD-VAE at the latent space.


\section{Proposed Class-Specific VAE}\label{sec:csvae}
As described above, the class-specific image retrieval task is defined as the task of retrieving images based on a query image which either belongs to a class of interest, or
not. One could approach this problem by applying the RD-VAE method described above, i.e., to consider the class of interest as the positive class and form a negative class including all images belonging to all other classes (binary RD-VAE, Figure \ref{Losses}-b). 
Since all VAE models determine the data representations in the latent space through training, one would assume that (using an adequately high number of network parameters and extensively tuning them) it is possible to force all training images forming the negative class (despite their possibly high variations) to form a homogeneous cluster in the latent space. However, this approach leads to increasing the complexity of the model and may not be able to generalize well on unseen (test) data. 

The Class-Specific VAE (CS-VAE) introduces a new KLD term and a new repulsive term in the training loss of the VAE:
\begin{equation}
L_{\operatorname{cs-vae}}\!=\!MSE_{loss}\!+\!\alpha_{\rm KL}KLD_{sup}^{cs}\!+\!rep_{loss}^{cs},\label{L_cs}
\end{equation}
\begin{equation}
KLD_{sup}^{cs}\!=\!\\
\mathlarger{\sum_{x_i\in l_p}}{\left({(\,x_i\!\!-\mu_{l_p})\,}\!^2\!+\!\sigma_{l_p}\!^2\!-\!\log(\,\sigma_{l_p})\,\!-\!1\right)}, \label{KLD_s_cs}
\end{equation}
\begin{equation}
rep_{loss}^{cs}\!=\!\sfrac{1}{\rho}\mathlarger{\sum_{\mathclap{x_i\notin l_p}}}{\max\left(0,\rho \!-\!\parallel x_{i}\!-\!\mu_{l_p} \parallel_{2}^{2}\right)\!^2}.\label{Rep_cs}
\end{equation}
Optimizing the loss function in Eq. (\ref{L_cs}) forces the image representations in the latent space of the positive class to form a Gaussian distribution, defined by mean $\mu_{l_p}$ and variance $\sigma_{l_p}$. Moreover, the image representations in the latent space of the data belonging to the negative class are forced to be far away (with a minimum distance of $\rho$) from the mean of the positive class. Those data representations are also constrained by the $MSE_{loss}$ loss, meaning that they need to preserve adequate input information in order to reconstruct the input image in the output of the Decoder. 

Figure \ref{Losses}-c shows a schematic 2D representation of CS-VAE at the latent space. As can be seen in that Figure, one would expect that such image representations in the latent space can have some favourable properties for class-specific CBIR. Representations of images belonging to the class of interest (positive class) are learnt to lay close to each other in the latent space (i.e., to be similar to each other, meaning that properties of those images in common are expected to be highlighted). Representations of images not belonging to the class of interest are forced to be well-discriminated in relation to the class of interest and they are not forced to group together, i.e., they are allowed to lay anywhere in the latent space leading to representations of dissimilar input images of the negative class to be far away from each other. This can lead to better-preserving properties of images belonging to the negative class and higher performance. 

It should be noted that class-specific optimization criteria have also been used in the past for determining data representations based on linear \cite{tran2017multilinear, iosifidis2018probabilistic} and kernel-based \cite{goudelis2007classspecific, iosifidis2015CSRDA, iosifidis2017cskdaRev, li2022classspecific} Class-Specific Discriminant Analysis. Those methods commonly optimize a class-specific variant of the Rayleigh Quotient \cite{martinez2001pca, wang2007trace} which tries to minimize the in-class scatter to out-of-class scatter ratio in the projection space. Contrary to this approach, the proposed CS-VAE exploits the KLD and the repulsive terms in Eqs. (\ref{KLD_s_cs}) and (\ref{Rep_cs}) which, in combination to the reconstruction error at the output of the Decoder, lead to the optimization problem in Eq. (\ref{L_cs}) which is well-suited for training deep learning models. 


\section{Experiments}\label{sec:Imp}
\begin{figure}[]
\centering
\includegraphics[width=\columnwidth]{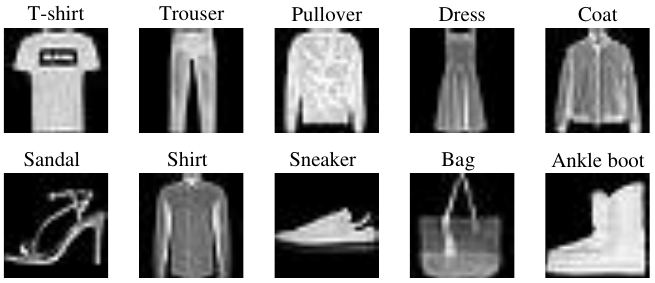}
\caption{Example images from the Fashion MNIST dataset.}
\label{FashinMNIST}
\end{figure}

\begin{figure}[]
\centering
\includegraphics[width=\columnwidth]{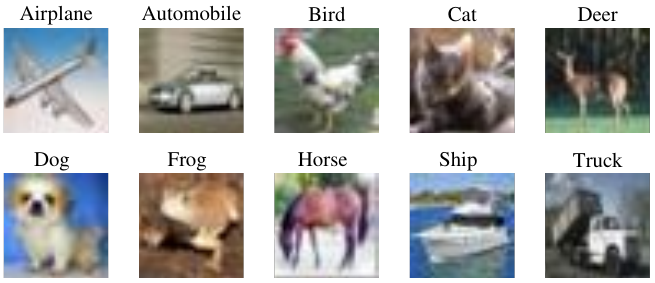}
\caption{Example images from the Cifar-10 dataset.}
\label{Cifar-10}
\end{figure}

\begin{figure}[]
\centering
\includegraphics[width=\columnwidth]{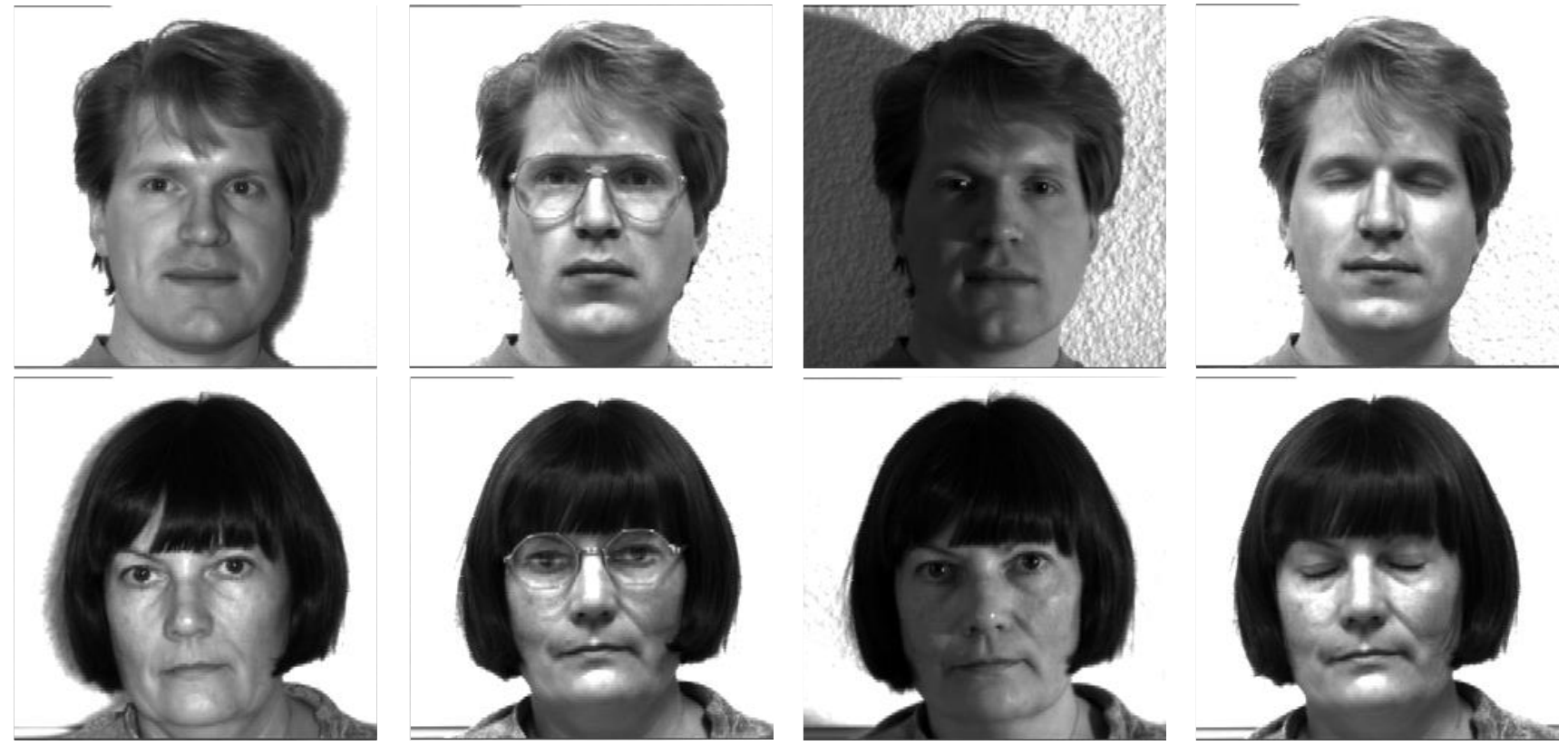}
\caption{Example images from the Yale dataset.}
\label{Yale}
\end{figure}

To evaluate the performance of the proposed CS-VAE method, we conducted experiments on four datasets. We used the following three publicly available datasets:
\begin{itemize}
    \item \textit{Fashion MNIST} \cite{xiao2017fashion}: It includes 70,000 fashion-related gray-scale images with resolution of $28\times28$ pixels. The images belong to 10 classes, with 7,000 images per class. 60,000 images form the training set and the remaining 10,000 the test set. Figure \ref{FashinMNIST} shows example images from the dataset. 
    
    \item \textit{Cifar-10} \cite{krizhevsky2009learning}: It includes 60,000 RGB-color images with resolution of $32\times32$ pixels. The images belong to 10 classes, with 6,000 images per class. 50,000 images form the training set and the remaining 10,000 the test set. Figure \ref{Cifar-10} shows example images from the dataset. 
    
    \item \textit{Yale} \cite{belhumeur1996eigenfaces}: It includes 165 gray-scale facial images of 15 different subjects (11 images per subject). The images have $320\times243$ pixels resolution and are taken under different lighting and facial expressions. Figure \ref{Yale} shows example images from the dataset.

\end{itemize}

To also evaluate the performance of the proposed method on a problem coming from an industrial application, we also used an X-ray image dataset of fibrous products. This dataset was collected over an X-ray test on several defective and non-defective fibrous product samples. The dataset contains four classes including a Non-defective (ND) class and the following three defective classes:
\begin{itemize}
    \item \textbf{D1:} the drops of melted raw materials that are not converted to fibers successfully;
    
    \item \textbf{D2:} binder bulks that are not evenly distributed over the fibers;
     
    \item \textbf{D3:} a collection of several small shots of molten raw materials close together.
\end{itemize}

\begin{figure}[]
\centering
\includegraphics[width=\columnwidth]{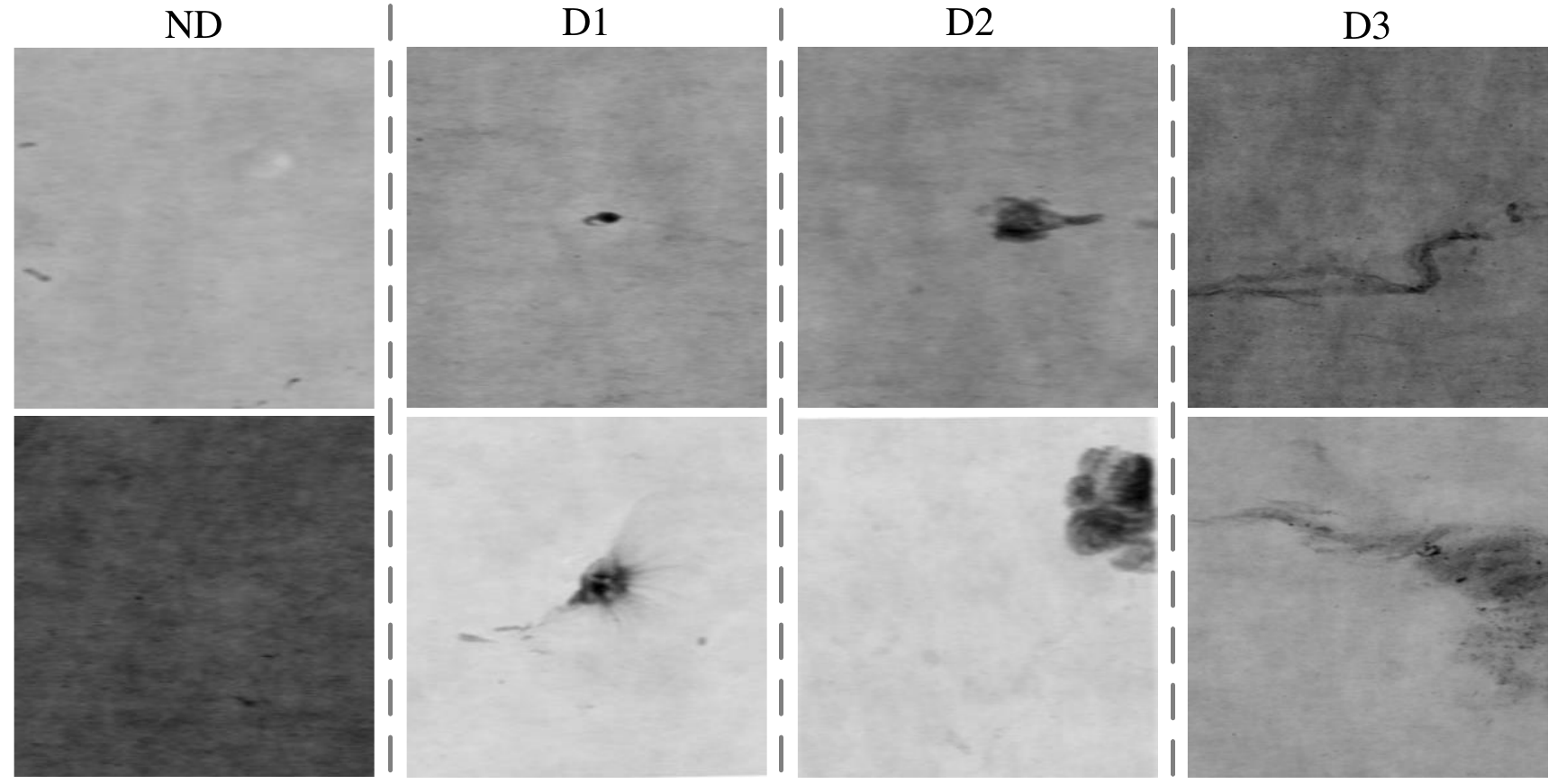}
\caption{Example images from the X-ray dataset.}
\label{X-ray}
\end{figure}

Samples of these four classes are shown in Figure \ref{X-ray}. 
In total, 271 gray-scale images with a resolution of $244\times244$ pixels are available in this dataset. In our experiments, only the ND class is considered as the class of interest, since this approach resembles a real-life anomaly detection problem where the defects need to be distinguished from the non-defective class in order to retrieve the non-defective products.

\subsection{In-domain CBIR experiments}

\begin{figure*}[]
\centering
\includegraphics[width=.9\textwidth]{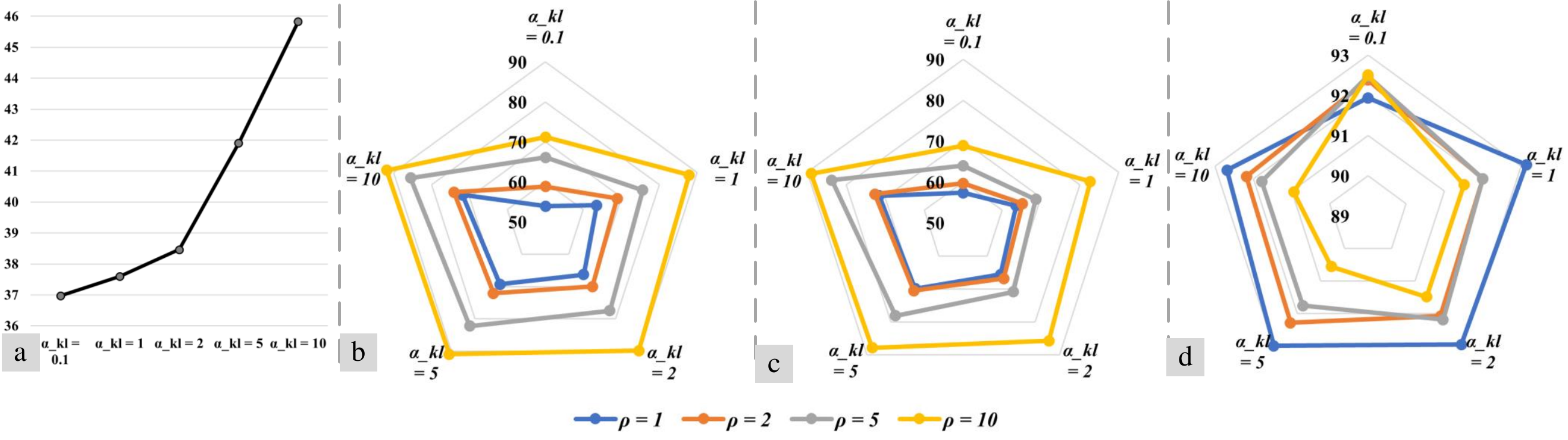}
\caption{Performance (mAP\%) on Fashion MNIST (in-domain retrieval) for different values of $\rho$ and $\alpha_{\rm KL}$: a) VAE, b) RD-VAE, c) binary RD-VAE, and d) CS-VAE}
\label{hyper_fig}
\end{figure*}
 
To evaluate the performance of the proposed method in CBIR based on query images belonging to classes included in the training, we conducted experiments following the in-domain experimental protocol. We compare the performance of CS-VAE with that of three other VAE-based methods, i.e., VAE, RD-VAE, and binary RD-VAE. Since VAE and RD-VAE methods are trained based on unsupervised and multi-class optimization problems, respectively, one model is trained on each dataset and the 11-recall point-based Average Precision (AP) metric \citep{manning2008introduction} is calculated for each class separately, as well as the mean AP (mAP) over all classes. For Binary RD-VAE and CS-VAE, a model is trained for each class-specific problem by considering the corresponding class as the class of interest, and AP is calculated for that class. We also calculated the mAP of all class-specific models. We repeated the experiments five times and we reported the mean and standard deviation of AP values for all experiments.

For hyper-parameter selection on the experiments in Fashion MNIST and Cifar-10 datasets, the training set is randomly split into 80\%/20\% training/validation subsets, and the values of the hyper-parameters of all models are selected based on their performance on the validation set. Due to the small number of samples per class in Yale and the X-ray datasets, we performed $5$-fold cross-validation, where the performance of each model having different hyper-parameter values is evaluated as the average AP over all folds.

Considering Equations \eqref{L_vae} - \eqref{Rep_cs}, hyper-parameter selection is applied to select values for $\rho$ and $\alpha_{\rm KL}$, and for determining the model architecture. We use a grid search strategy to select the values of $\rho$ and $\alpha_{\rm KL}$ for each dataset and each model using the ranges $\rho=\{1,\dots,10\}$ and $\alpha_{\rm KL} = \{0.1,\dots,10\}$. For instance, the mAP values over all classes on the Fashion MNIST dataset are shown in Figure \ref{hyper_fig}. It can be seen from Figure \ref{hyper_fig}-a, b, and c that there is a linear relation between the hyper-parameters value and the models' performance. However, for CS-VAE, the optimal values of these two hyper-parameters are in the range of the selected intervals.

The selected model parameters, including the model's input size, encoder and decoder architecture, the size of latent space, and the two hyper-parameter values ($\rho$ and $\alpha_{\rm KL}$), for all datasets and methods are shown in Table \ref{hyper-par-table}. It can be seen that to create the models' encoders, 2D convolution layers, batch normalization layers, and ReLU activation functions are used, followed by flattening and two parallel linear layers to have means and variances for each dimension of the latent space. For the decoders, after linear and unflattening layers, 2D transposed convolution layers, batch normalization layers, and ReLU activation functions are used, followed by a sigmoid activation function to construct the output image.

\input{tables/hyper.tex}

\noindent
\textit{\textbf{Results:}}
The performance (AP\%) of each method on each class of Fashion-MNIST, as well as the mAP over all classes, are reported in Table \ref{in-domain_fmnist_results}. From this table, it can be seen that RD-VAE achieved much higher performance in comparison to the VAE model. Such an improvement was expected due to the separate Gaussian distributions defined by RD-VAE for each class and the use of the repulsive loss to push them away from each other in the latent space. The binary RD-VAE reached a slightly lower performance compared to RD-VAE. As it was explained in Section \ref{sec:csvae}, such a lower performance is expected as a result of the model forcing all negative image representations to form a homogeneous cluster in the latent space, despite their possibly high variations in the input space. Finally, CS-VAE managed to achieve the highest performance. 

\begin{figure}[]
\centering
\includegraphics[width=\columnwidth]{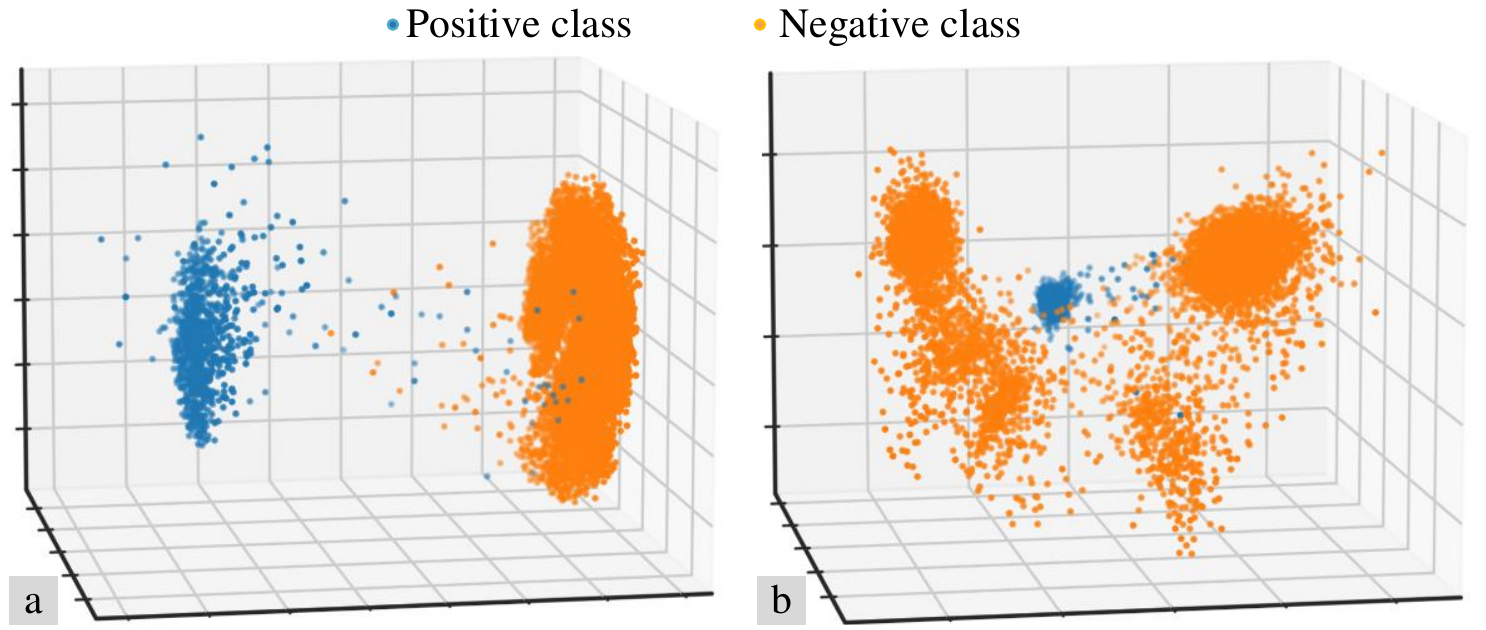}
\caption{Image representations of Fashion MNIST obtained by applying PCA on the latent space (three eigenvectors): a) binary RD-VAE, and b) CS-VAE}
\label{3D_pca}
\end{figure}

Figures \ref{3D_pca}-a and b illustrate the data representation obtained by applying Principal Component Analysis on the image representations in the latent spaces of the binary RD-VAE and CS-VAE and keeping the top three eigenvectors. It can be seen that for the binary RD-VAE, all samples belonging to the negative class are pushed to cluster one side of the positive class. CS-VAE, as also mentioned in Section \ref{sec:csvae}, allows the representations of the negative images in the latent space to freely be arranged, as long as they are adequately far away (parameterized by the value of $\rho$) from the positive class' mean.

\input{tables/in_fmnist.tex}

The performance of all competing methods on the Cifar-10 dataset is reported in Table \ref{in-domain_cifar-10_results}. As can be seen, similar observations to those made for the results on Fashion MNIST can be made here too.
 
\input{tables/in_cifar.tex}

By observing the results reported for the Yale dataset in Table \ref{in-domain_yale_results} it can be seen that, although CS-VAE still achieves the highest performance, the binary RD-VAE outperforms the RD-VAE method. This can be due to lower variations in the images belonging to the negative class.
 
\input{tables/in_yale.tex}

Experimental results on the X-ray dataset are reported in Table \ref{in-domain_x-ray_results}. As described in Section \ref{sec:Imp}, on this dataset only the ND class is considered as the class of interest and, therefore, only one model is trained for each method and the AP\%s values of all methods are reported. Similarly to Fashion MNIST and Cifar-10, CS-VAE achieved the highest performance followed by RD-VAE.
 
\input{tables/in_x.tex}

\subsection{Out-of-domain CBIR experiments}
To evaluate the performance of the proposed method in retrieving images belonging to classes that are not present in the training phase, we also conducted out-of-domain experiments. To do this, only half of the available classes in the dataset are used to form the negative class in the training phase, which is selected randomly. For testing, we used all available classes in the dataset to form the negative class. Therefore, before splitting the training data into training and validation sets, half of the classes (excluding the class of interest) are randomly discarded. Then, the same experimental protocols and hyper-parameter selection processes as in the in-domain experiments are used. The selected hyper-parameters are shown in Table \ref{hyper-par-table}.

\noindent
\textit{\textbf{Results:}} The performance of binary RD-VAE and CS-VAE on all datasets is reported in Table \ref{out-domain_results}. In all cases, we see an accuracy drop compared to the in-domain experiments. However, this performance drop is higher for the binary RD-VAE. This shows that the proposed CS-VAE is more capable in learning better latent space representations for data belonging in unseen during training classes.

\input{tables/out.tex}

\section{Conclusion}\label{sec:conclusion}

In this paper, a variant of the VAE was proposed which is suited for class-specific content-based image retrieval. The proposed method models the optimization problem of the VAE to determine a latent space in which the class of interest is well-discriminated by samples belonging to any other class. To do this and for preserving as much information as possible in order to achieve a good reconstruction in its output, the model learns a discriminative data representation using KLD and repulsive losses forcing the data belonging to the class of interest to form a Gaussian distribution, while forcing all other samples far away from the mean of this distribution. This method was extensively evaluated over several public and custom datasets on both in-domain and out-of-domain retrieval tasks. Three related VAE-based methods were used for comparisons and the proposed method outperformed them in all cases.

\section*{Acknowledgment}
\noindent
The research leading to the results of this paper received funding from the Innovation Fund Denmark as part of MADE FAST.

\bibliographystyle{ieeetr}
\bibliography{IJCNN_CS-VAE}

\end{document}

%% file: tables/hyper.tex
\begin{table*}[]
 \caption{\label{hyper-par-table} Model and selected hyperparameter values for each method and dataset.}
 \centering
 \footnotesize
 \resizebox{\textwidth}{!}{
 \begin{tabular}{p{11mm}llll}
 \hline
\rowcolor[HTML]{808080} 
   & Fashion MNIST   & CIFAR-10  & Yale   & X-ray  \\ \hline
\cellcolor[HTML]{808080}Model & \begin{tabular}[t]{@{}l@{}}\textbf{Input:} 1$\times$28$\times$28\\ \hline \textbf{Encoder:}\\ C2d(1, 16, k=3,   s=1)\\ BN2d(16), ReLU()\\ C2d(16, 32,   k=3, s=2)\\ BN2d(32), ReLU()\\ C2d(32, 64,   k=3, s=1)\\ BN2d(64), ReLU()\\ C2d(64,   128, k=3, s=2)\\ BN2d(128), ReLU()\\ Flatten()\\ Linear(in=2048,   out=256)\\ BN1d(256), ReLU()\\ Linear(in=256,   out=30)\\ Linear(in=256,   out=30)\\ \hline \textbf{Decoder:}\\ Linear(in=30,   out=256)\\ BN1d(256), ReLU()\\ Linear(in=256,   out=2048)\\ BN1d(2048),,   ReLU()\\ UnFlatten()\\ CT2d(128,   64, k=3, s=3)\\ BN2d(64), ReLU()\\ CT2d(64,   32, k=3, s=2)\\ BN2d(32), ReLU()\\ CT2d(32,   16, k=3, s=1)\\ BN2d(16), ReLU()\\ CT2d(16, 1,   k=2, s=1)\\ Sigmoid()\end{tabular} & \begin{tabular}[t]{@{}l@{}}\textbf{Input:} 3$\times$32$\times$32\\ \hline \textbf{Encoder:}\\ C2d(3, 32, k=3,   s=1)\\ BN2d(32),   ReLU()\\ C2d(32, 64,   k=3, s=2)\\ BN2d(64),   ReLU()\\ C2d(64,   128, k=3, s=1)\\ BN2d(128), ReLU()\\ C2d(128,   256, k=3, s=2)\\ BN2d(256), ReLU()\\ Flatten()\\ Linear(in=6400,   out=1024)\\ BN1d(1024),   ReLU()\\ Linear(in=1024,   out=30)\\ Linear(in=1024,   out=30)\\ \hline \textbf{Decoder:}\\ Linear(in=30,   out=1024)\\ BN1d(1024),   ReLU()\\ Linear(in=1024,   out=6400)\\ BN1d(6400),   ReLU()\\ UnFlatten()\\ CT2d(256,   128, k=3, s=2)\\ BN2d(128), ReLU()\\ CT2d(128,   64, k=3, s=1)\\ BN2d(64),   ReLU()\\ CT2d(64,   32, k=3, s=2)\\ BN2d(32),   ReLU()\\ CT2d(32,   16, k=3, s=1)\\ BN2d(16), ReLU()\\ CT2d(16, 3,   k=(4, 4), s=1)\\ Sigmoid()\end{tabular} & \begin{tabular}[t]{@{}l@{}}\textbf{Input:} 1$\times$244$\times$244\\ \hline \textbf{Encoder:}\\ C2d(1, 32,   k=3, s=3)\\ BN2d(32), ReLU()\\ C2d(32, 64,   k=3, s=3)\\ BN2d(64), ReLU()\\ C2d(64, 128,   k=3, s=3)\\ BN2d(128), ReLU()\\ Flatten()\\ Linear(in=10368,   out=256)\\ BN1d(256), ReLU()\\ Linear(in=256,   out=30)\\ Linear(in=256,   out=30)\\ \\ \\ \hline \textbf{Decoder:}\\ Linear(in=30,   out=256)\\ BN1d(256), ReLU()\\ Linear(in=256,   out=10368)\\ BN1d(10368),   ReLU()\\ UnFlatten()\\ CT2d(128,   64, k=3, s=3)\\ BN2d(64), ReLU()\\ CT2d(64,   32, k=3, s=3)\\ BN2d(32), ReLU()\\ CT2d(32, 1,   k=4, s=3)\\ Sigmoid()\end{tabular} & \begin{tabular}[t]{@{}l@{}}\textbf{Input:} 1$\times$244$\times$244\\ \hline \textbf{Encoder:}\\ C2d(1, 32,   k=3, s=3)\\ BN2d(32), ReLU()\\ C2d(32, 64,   k=3, s=3)\\ BN2d(64), ReLU()\\ C2d(64, 128,   k=3, s=3)\\ BN2d(128), ReLU()\\ Flatten()\\ Linear(in=10368,   out=256)\\ BN1d(256), ReLU()\\ Linear(in=256,   out=10)\\ Linear(in=256,   out=10)\\ \\ \\ \hline \textbf{Decoder:}\\ Linear(in=10,   out=256)\\ BN1d(256), ReLU()\\ Linear(in=256,   out=10368)\\ BN1d(10368),   ReLU()\\ UnFlatten()\\ CT2d(128,   64, k=3, s=3)\\ BN2d(64), ReLU()\\ CT2d(64,   32, k=3, s=3)\\ BN2d(32), ReLU()\\ CT2d(32, 1,   k=4, s=3)\\ Sigmoid()\end{tabular} \\ \hline

\cellcolor[HTML]{808080}VAE   & $\alpha_{kl}=10$   & $\alpha_{kl}=10$   & $\alpha_{kl}=5$ & $\alpha_{kl}=2$ \\ \hline

\cellcolor[HTML]{808080} RD-VAE  & \begin{tabular}[]{@{}l@{}} $\rho=10$  \\ \hline $\alpha_{kl}=10$ \end{tabular} & \begin{tabular}[]{@{}l@{}} $\rho=10$  \\ \hline $\alpha_{kl}=10$ \end{tabular} & \begin{tabular}[]{@{}l@{}} $\rho=10$  \\ \hline $\alpha_{kl}=10$ \end{tabular} & \begin{tabular}[]{@{}l@{}} $\rho10=$  \\ \hline $\alpha_{kl}=10$ \end{tabular} \\ \hline

\cellcolor[HTML]{808080} \begin{tabular}[c]{@{}l@{}}Binary \\ RD-VAE\end{tabular}  & \begin{tabular}[]{@{}l@{}} \begin{tabular}[c]{@{}l@{}}In-domain: $\rho=10$\\ Out-of-domain:   $\rho=10$\end{tabular} \\ \hline \begin{tabular}[c]{@{}l@{}}In-domain: $\alpha_{kl}=10$\\ Out-of-domain: $\alpha_{kl}=10$\end{tabular} \end{tabular} & \begin{tabular}[]{@{}l@{}} \begin{tabular}[c]{@{}l@{}}In-domain: $\rho=10$\\ Out-of-domain:   $\rho=10$\end{tabular} \\ \hline \begin{tabular}[c]{@{}l@{}}In-domain: $\alpha_{kl}=10$\\ Out-of-domain: $\alpha_{kl}=10$\end{tabular} \end{tabular} & \begin{tabular}[]{@{}l@{}} \begin{tabular}[c]{@{}l@{}}In-domain: $\rho=10$\\ Out-of-domain:   $\rho=10$\end{tabular} \\ \hline \begin{tabular}[c]{@{}l@{}}In-domain: $\alpha_{kl}=10$\\ Out-of-domain: $\alpha_{kl}=10$\end{tabular} \end{tabular} & \begin{tabular}[]{@{}l@{}} \begin{tabular}[c]{@{}l@{}}In-domain: $\rho=10$\\ Out-of-domain:   $\rho=10$\end{tabular} \\ \hline \begin{tabular}[c]{@{}l@{}}In-domain: $\alpha_{kl}=10$\\ Out-of-domain: $\alpha_{kl}=10$\end{tabular} \end{tabular} \\  \hline

\cellcolor[HTML]{808080} CS-VAE  & \begin{tabular}[]{@{}l@{}} \begin{tabular}[c]{@{}l@{}}In-domain: $\rho=1$\\ Out-of-domain:   $\rho=1$\end{tabular} \\ \hline \begin{tabular}[c]{@{}l@{}}In-domain: $\alpha_{kl}=1$\\ Out-of-domain: $\alpha_{kl}=2$\end{tabular} \end{tabular} & \begin{tabular}[]{@{}l@{}} \begin{tabular}[c]{@{}l@{}}In-domain: $\rho=2$\\ Out-of-domain:   $\rho=5$\end{tabular} \\ \hline \begin{tabular}[c]{@{}l@{}}In-domain: $\alpha_{kl}=10$\\ Out-of-domain: $\alpha_{kl}=2$\end{tabular} \end{tabular} & \begin{tabular}[]{@{}l@{}} \begin{tabular}[c]{@{}l@{}}In-domain: $\rho=10$\\ Out-of-domain:   $\rho=10$\end{tabular} \\ \hline \begin{tabular}[c]{@{}l@{}}In-domain: $\alpha_{kl}=5$\\ Out-of-domain: $\alpha_{kl}=5$\end{tabular} \end{tabular} & \begin{tabular}[]{@{}l@{}} \begin{tabular}[c]{@{}l@{}}In-domain: $\rho=10$\\ Out-of-domain:   $\rho=5$\end{tabular} \\ \hline \begin{tabular}[c]{@{}l@{}}In-domain: $\alpha_{kl}=5$\\ Out-of-domain: $\alpha_{kl}=10$\end{tabular} \end{tabular}  \\  \hline

 \end{tabular}}

 \vspace{1mm}
 
 \footnotesize
 C2d: 2D convolution, CT2d: 2D transposed convolution, BN2d: 2D Batch Normalization, BN1d: 1D Batch Normalization.
 
 \end{table*}

%% file: tables/in_fmnist.tex
\begin{table}[]
 \caption{\label{in-domain_fmnist_results} Evaluation results on Fashion MNIST dataset for in-domain retrieval.}
 \centering
 \begin{tabular}{lcccc}
 \hline
 \rowcolor[HTML]{808080} 
 \cellcolor[HTML]{808080} & \multicolumn{4}{c}{\cellcolor[HTML]{808080}AP\% (mean± std)} \\ \cline{2-5} 
 \rowcolor[HTML]{808080} 
 \multirow{-2}{*}{\cellcolor[HTML]{808080}} & VAE & RD-VAE & Binary RD-VAE & CS-VAE \\ \hline
 C1 & 48.48±0.33 & 89.32±0.46 & 82.77±2.10 & \textbf{89.94±1.08} \\ \hline
 C2 & 65.14±1.61 & 96.78±0.19 & 96.55±2.34 & \textbf{97.50±0.32} \\ \hline
 C3 & 34.96±0.88 & \textbf{89.20±0.66} & 83.72±1.28 & 89.14±1.25 \\ \hline
 C4 & 42.22±0.42 & 91.22±0.35 & 87.87±0.53 & \textbf{92.72±0.39} \\ \hline
 C5 & 36.20±0.41 & 87.48±0.18 & 81.02±0.52 & \textbf{90.16±0.84} \\ \hline
 C6 & 47.04±2.56 & 96.86±0.29 & 97.67±0.23 & \textbf{98.02±0.11} \\ \hline
 C7 & 28.68±0.22 & 76.78±0.55 & 68.15±0.82 & \textbf{82.22±0.74} \\ \hline
 C8 & 56.50±0.99 & 96.62±0.30 & 97.37±0.17 & \textbf{97.56±0.30} \\ \hline
 C9 & 45.08±3.31 & 96.82±0.07 & \textbf{97.50±0.23} & 97.44±0.23 \\ \hline
 C10 & 54.08±2.51 & 96.52±0.19 & 96.45±0.25 & \textbf{96.76±0.16} \\ \hline
 \rowcolor[HTML]{D9D9D9} 
 Mean & 45.83 & 91.75 & 89.02 & \textbf{93.14} \\ \hline
 \end{tabular}
 \end{table}

%% file: tables/in_cifar.tex
\begin{table}[]
 \caption{\label{in-domain_cifar-10_results} Evaluation results on Cifar-10 dataset for in-domain retrieval.}
 \centering
 \begin{tabular}{lcccc}
 \hline
 \rowcolor[HTML]{808080} 
 \cellcolor[HTML]{808080} & \multicolumn{4}{c}{\cellcolor[HTML]{808080}AP\% (mean± std)} \\ \cline{2-5} 
 \rowcolor[HTML]{808080} 
 \multirow{-2}{*}{\cellcolor[HTML]{808080}} & VAE  & RD-VAE  & Binary RD-VAE & CS-VAE  \\ \hline
C1  & 19.10±0.32 & 71.62±0.55 & 59.57±2.35 & \textbf{76.95±3.48} \\ \hline
C2  & 15.17±0.21 & 77.77±0.87 & 68.47±3.85 & \textbf{83.92±2.98} \\ \hline
C3  & 17.07±0.16 & 59.67±0.53 & 45.77±0.31 & \textbf{63.07±3.52} \\ \hline
C4  & 13.50±0.12 & 53.07±0.72 & 40.12±1.14 & \textbf{58.95±0.26} \\ \hline
C5  & 18.62±0.13 & 63.05±0.68 & 49.95±0.81 & \textbf{69.22±3.01} \\ \hline
C6  & 14.77±0.19 & 62.10±0.50 & 47.42±0.92 & \textbf{66.57±2.41} \\ \hline
C7  & 19.90±0.25 & 74.12±0.10 & 67.17±1.97 & \textbf{81.17±2.99} \\ \hline
C8  & 15.50±0.14 & 69.57±1.09 & 61.87±0.86 & \textbf{77.00±1.70} \\ \hline
C9  & 19.45±0.27 & 78.32±0.95 & 71.07±0.92 & \textbf{85.75±2.15} \\ \hline
C10 & 16.22±0.14 & 75.15±1.34 & 62.47±3.67 & \textbf{81.32±1.86} \\ \hline
\rowcolor[HTML]{D9D9D9} 
Mean   & 16.93   & 68.44   & 57.39   & \textbf{74.39}   \\ \hline
 \end{tabular}
 \end{table}

%% file: tables/in_yale.tex
\begin{table}[]
 \caption{\label{in-domain_yale_results} Evaluation results on Yale dataset for in-domain retrieval.}
 \centering
 \begin{tabular}{lllll}
 \hline
 \rowcolor[HTML]{808080} 
 \cellcolor[HTML]{808080} & \multicolumn{4}{c}{\cellcolor[HTML]{808080}AP\% (mean± std)} \\ \cline{2-5} 
 \rowcolor[HTML]{808080} 
 \multirow{-2}{*}{\cellcolor[HTML]{808080}} & VAE  & RD-VAE  & Binary RD-VAE  & CS-VAE  \\ \hline
C1  & 72.20±1.75 & 98.21±1.98 & 99.12±1.09  & \textbf{99.20±1.16} \\ \hline
C2  & 74.40±0.74 & 98.18±1.39 & 98.92±0.92  & \textbf{99.20±0.91} \\ \hline
C3  & 74.46±1.20 & 99.52±2.47 & \textbf{100.0±0.00}  & \textbf{100.0±0.00} \\ \hline
C4  & 73.32±1.49 & 99.33±0.38 & \textbf{99.94±0.08}  & 99.84±0.16 \\ \hline
C5  & 73.50±1.11 & 99.21±1.08 & \textbf{99.98±0.04}  & 99.96±0.08 \\ \hline
C6  & 73.45±0.98 & 99.03±0.84 & 99.52±0.66  & \textbf{99.94±0.04} \\ \hline
C7  & 72.68±0.74 & 98.97±1.89 & \textbf{99.56±0.34}  & 98.48±1.57 \\ \hline
C8  & 74.45±1.97 & 98.12±0.33 & 98.90±1.24  & \textbf{99.64±0.53} \\ \hline
C9  & 76.62±2.05 & 98.30±1.60 & 98.44±1.48  & \textbf{99.64±0.43} \\ \hline
C10 & 75.91±1.98 & 98.52±0.67 & 99.78±0.34  & \textbf{99.96±0.08} \\ \hline
\rowcolor[HTML]{FFFFFF} 
C11 & 77.01±1.40 & 99.35±0.81 & \textbf{99.72±0.24}  & 99.06±1.17 \\ \hline
C12 & 77.21±0.70 & 99.01±0.32 & 99.34±1.12  & \textbf{99.90±0.15} \\ \hline
C13 & 72.23±0.41 & 99.15±1.09 & \textbf{99.82±0.14}  & 98.70±1.08 \\ \hline
C14 & 74.05±0.76 & 98.63±0.84 & 98.10±1.90  & \textbf{99.44±0.30} \\ \hline
C15 & 73.05±1.12 & 99.24±1.60 & \textbf{99.50±0.63}  & 99.00±1.42 \\ \hline
\rowcolor[HTML]{D9D9D9} 
Mean   & 74.30   & 98.85   & \cellcolor[HTML]{D9D9D9}99.37 & \textbf{99.46}   \\ \hline
 \end{tabular}
 \end{table}

%% file: tables/in_x.tex
\begin{table}[]
 \caption{\label{in-domain_x-ray_results} Evaluation results on X-ray dataset for in-domain implementation.}
 \centering
 \resizebox{\columnwidth}{!}{
 \begin{tabular}{lcccc}
 \hline
 \rowcolor[HTML]{808080} 
 \cellcolor[HTML]{808080} & \multicolumn{4}{c}{\cellcolor[HTML]{808080}AP\% (mean± std)} \\ \cline{2-5} 
 \rowcolor[HTML]{808080} 
 \multirow{-2}{*}{\cellcolor[HTML]{808080}} & VAE & RD-VAE & Binary RD-VAE & CS-VAE \\ \hline
 ND & 39.05±0.54 & 75.27±3.03 & 68.30±1.20 & \textbf{94.50±1.69} \\ \hline
 \end{tabular}}
 \end{table}

%% file: tables/out.tex
\begin{table}[]
 \caption{\label{out-domain_results} Evaluation results for out-domain implementation.}
 \centering
 \begin{tabular}{
 >{\columncolor[HTML]{808080}}l cc}
\hline
\cellcolor[HTML]{808080} & \multicolumn{2}{c}{\cellcolor[HTML]{808080} mAP\%}  \\ \cline{2-3} 
\multirow{-2}{*}{\cellcolor[HTML]{808080}} & \cellcolor[HTML]{808080}Binary RD-VAE & \cellcolor[HTML]{808080}CS-VAE \\ \hline
Fashion MNIST   & 71.39   & \textbf{80.19}  \\ \hline
Cifar-10  & 54.02   & \textbf{69.32}  \\ \hline
Yale   & 97.85   & \textbf{98.59}  \\ \hline
X-ray  & 71.55   & \textbf{93.85}  \\ \hline
 \end{tabular}
 \end{table}

%% file: IJCNN_CS-VAE.bbl
\begin{thebibliography}{10}

\bibitem{chen2022deep}
W.~Chen, Y.~Liu, W.~Wang, E.~M. Bakker, T.~Georgiou, P.~Fieguth, L.~Liu, and
  M.~S. Lew, ``Deep learning for instance retrieval: A survey,'' {\em IEEE
  Transactions on Pattern Analysis and Machine Intelligence}, pp.~1--20, 2022.

\bibitem{alam2020dynamic}
K.~M. Alam, N.~Siddique, and H.~Adeli, ``A dynamic ensemble learning algorithm
  for neural networks,'' {\em Neural Computing and Applications}, vol.~32,
  no.~12, pp.~8675--8690, 2020.

\bibitem{liu2013content}
G.-H. Liu and J.-Y. Yang, ``Content-based image retrieval using color
  difference histogram,'' {\em Pattern Recognition}, vol.~46, no.~1,
  pp.~188--198, 2013.

\bibitem{singha2012content}
M.~Singha and K.~Hemachandran, ``Content based image retrieval using color and
  texture,'' {\em Signal \& Image Processing: An International Journal},
  vol.~3, no.~1, pp.~39--57, 2012.

\bibitem{lowe2004distinctive}
D.~G. Lowe, ``Distinctive image features from scale-invariant keypoints,'' {\em
  International Journal of Computer Vision}, vol.~60, no.~2, pp.~91--110, 2004.

\bibitem{sivic2003video}
J.~Sivic and A.~Zisserman, ``Video google: A text retrieval approach to object
  matching in videos,'' {\em IEEE International Conference on Computer Vision},
  vol.~2, pp.~1470--1477, 2003.

\bibitem{uchida2016image}
Y.~Uchida, S.~Sakazawa, and S.~Satoh, ``Image retrieval with fisher vectors of
  binary features,'' {\em ITE Transactions on Media Technology and
  Applications}, vol.~4, no.~4, pp.~326--336, 2016.

\bibitem{pan2022generating}
J.~Pan, X.~Zhu, and P.~Liu, ``Generating adaptive targeted adversarial examples
  for content-based image retrieval,'' {\em International Joint Conference on
  Neural Networks}, pp.~1--9, 2022.

\bibitem{hamreras2021dynamic}
S.~Hamreras, B.~Boucheham, M.~A. Molina-Cabello, R.~Ben{\'\i}tez-Rochel, and
  E.~L{\'o}pez-Rubio, ``Dynamic selection of classifiers for content based
  image retrieval,'' {\em International Joint Conference on Neural Networks},
  pp.~1--8, 2021.

\bibitem{pang2018deep}
S.~Pang, J.~Ma, J.~Xue, J.~Zhu, and V.~Ordonez, ``Deep feature aggregation and
  image re-ranking with heat diffusion for image retrieval,'' {\em IEEE
  Transactions on Multimedia}, vol.~21, no.~6, pp.~1513--1523, 2018.

\bibitem{radenovic2018fine}
F.~Radenovi{\'c}, G.~Tolias, and O.~Chum, ``Fine-tuning cnn image retrieval
  with no human annotation,'' {\em IEEE Transactions on Pattern Analysis and
  Machine Intelligence}, vol.~41, no.~7, pp.~1655--1668, 2018.

\bibitem{deng2009imagenet}
J.~Deng, W.~Dong, R.~Socher, L.-J. Li, K.~Li, and L.~Fei-Fei, ``Imagenet: A
  large-scale hierarchical image database,'' {\em IEEE Conference on Computer
  Vision and Pattern Recognition}, pp.~248--255, 2009.

\bibitem{passalis2016entropy}
N.~Passalis and A.~Tefas, ``Entropy optimized feature-based bag-of-words
  representation for information retrieval,'' {\em IEEE Transactions on
  Knowledge and Data Engineering}, vol.~28, no.~7, pp.~1664--1677, 2016.

\bibitem{passalis2020variance}
N.~Passalis, A.~Iosifidis, M.~Gabbouj, and A.~Tefas, ``Variance-preserving deep
  metric learning for content-based image retrieval,'' {\em Pattern Recognition
  Letters}, vol.~131, pp.~8--14, 2020.

\bibitem{kingma2013auto}
D.~P. Kingma and M.~Welling, ``Auto-encoding variational bayes,'' {\em arXiv
  preprint arXiv:1312.6114}, 2013.

\bibitem{tran2017multilinear}
D.~T. Tran, M.~Gabbouj, and A.~Iosifidis, ``Multilinear class-specific
  discriminant analysis,'' {\em Pattern Recognition Letters}, vol.~100,
  pp.~131--136, 2017.

\bibitem{iosifidis2018probabilistic}
A.~Iosifidis, ``Probabilistic class-specific discriminant analysis,'' {\em IEEE
  Access}, pp.~183847--183855, 2020.

\bibitem{goudelis2007classspecific}
G.~Goudelis, S.~Zafeiriou, A.~Tefas, and I.~Pitas, ``Class-specific kernel
  discriminant analysis for face verification,'' {\em IEEE Transactions on
  Information Forensics and Security}, vol.~2, no.~3, pp.~570--587, 2007.

\bibitem{iosifidis2015CSRDA}
A.~Iosifidis, A.~Tefas, and I.~Pitas, ``Class-specific reference discriminant
  analysis with application in human behavior analysis,'' {\em IEEE
  Transactions on Human-Machine Systems}, vol.~45, no.~3, pp.~315--326, 2015.

\bibitem{iosifidis2017cskdaRev}
A.~Iosifidis and M.~Gabbouj, ``Class-specific kernel discriminant analysis
  revisited: Further analysis and extensions,'' {\em IEEE Transactions on
  Cybernetics}, vol.~47, no.~12, pp.~4485--4496, 2017.

\bibitem{li2022classspecific}
K.~Li and G.~Wu, ``Randomized approximate class-specific kernel spectral
  regression analysis for large-scale face verification,'' {\em Machine
  Learning}, vol.~111, no.~6, pp.~2037--2091, 2022.

\bibitem{martinez2001pca}
A.~Martinez and A.~Kak, ``{PCA} versus {LDA},'' {\em IEEE Transactions on
  Pattern Analysis and Machine Intelligence}, vol.~23, no.~2, pp.~228--233,
  2001.

\bibitem{wang2007trace}
H.~Wang, S.~Yan, D.~Xu, X.~Tang, and T.~Huang, ``Trace ratio vs. ratio trace
  for dimensionality reduction,'' {\em IEEE Conference on Computer Vision and
  Pattern Recognition}, pp.~1--8, 2007.

\bibitem{xiao2017fashion}
H.~Xiao, K.~Rasul, and R.~Vollgraf, ``Fashion-mnist: a novel image dataset for
  benchmarking machine learning algorithms,'' {\em arXiv preprint
  arXiv:1708.07747}, 2017.

\bibitem{krizhevsky2009learning}
A.~Krizhevsky and G.~Hinton, ``Learning multiple layers of features from tiny
  images,'' {\em Technical Report, Toronto, ON, Canada}, 2009.

\bibitem{belhumeur1996eigenfaces}
P.~N. Belhumeur, J.~P. Hespanha, and D.~J. Kriegman, ``Eigenfaces vs.
  fisherfaces: Recognition using class specific linear projection,'' {\em
  European Conference on Computer Vision}, vol.~19, no.~7, pp.~43--58, 1996.

\bibitem{manning2008introduction}
H.~Sch{\"u}tze, C.~D. Manning, and P.~Raghavan, ``Introduction to information
  retrieval,'' {\em Cambridge University Press}, vol.~39, pp.~234--265, 2008.

\end{thebibliography}
